\documentclass[sigconf]{acmart}

\usepackage{graphicx}
\usepackage{stfloats}
\usepackage{wrapfig}
\usepackage{lipsum}  

\newlength\savewidth
\newcommand\shline{\noalign{\global\savewidth\arrayrulewidth
                            \global\arrayrulewidth 1.5pt}%
                   \hline
                   \noalign{\global\arrayrulewidth\savewidth}
                   }

\usepackage{xcolor}

\usepackage{algorithm}
\usepackage[noend]{algpseudocode}

\algrenewcommand\alglinenumber[1]{#1}

\usepackage{multirow}

\usepackage{enumitem}
\setlist[itemize]{leftmargin=*, topsep=.0em, itemsep=0pt, parsep=0pt, partopsep=0pt}
\setlist[enumerate]{leftmargin=*, topsep=.0em, itemsep=0pt, parsep=0pt, partopsep=0pt}

\usepackage{amsthm}
\newtheoremstyle{def_style}
  {0.5em}      
  {0.5em}      
  {}          
  {}          
  {\bfseries} 
  {.}         
  {0.5em}      
  {}          

\theoremstyle{def_style}
\newtheorem{define}{Definition}

\theoremstyle{def_style}
\newtheorem{prob}{Problem}

\usepackage{amsmath}
\usepackage{bbm}

\usepackage{collectbox}



\usepackage{mathtools}
\newcommand{\defeq}{\vcentcolon=}

\newcommand{\iu}{{i\mkern1mu}}

\usepackage{balance}
\usepackage{pythonhighlight}

\copyrightyear{2024}
\acmYear{2024}
\setcopyright{acmlicensed}\acmConference[KDD '24]{Proceedings of the 30th
ACM SIGKDD Conference on Knowledge Discovery and Data Mining}{August
25--29, 2024}{Barcelona, Spain}
\acmBooktitle{Proceedings of the 30th ACM SIGKDD Conference on Knowledge
Discovery and Data Mining (KDD '24), August 25--29, 2024, Barcelona, Spain}
\acmDOI{10.1145/3637528.3671881}
\acmISBN{979-8-4007-0490-1/24/08}

\begin{document}

\newcommand{\proposed}{RPMixer}
\title[\proposed: Shaking Up Time Series Forecasting with Random Projections]{\proposed: Shaking Up Time Series Forecasting with Random Projections for Large Spatial-Temporal Data}













\author{Chin-Chia Michael Yeh}
\author{Yujie Fan}
\author{Xin Dai}
\author{Uday Singh Saini}
\affiliation{%
}

\author{Vivian Lai}
\author{Prince Osei Aboagye}
\author{Junpeng Wang}
\author{Huiyuan Chen}
\affiliation{%
  \institution{Visa Research}
  \state{California}
  \country{USA}
}

\author{Yan Zheng}
\author{Zhongfang Zhuang}
\author{Liang Wang}
\author{Wei Zhang}
\affiliation{%
}


\renewcommand{\shortauthors}{Chin-Chia Michael Yeh et al.}

\begin{abstract}
Spatial-temporal forecasting systems play a crucial role in addressing numerous real-world challenges. 
In this paper, we investigate the potential of addressing spatial-temporal forecasting problems using general time series forecasting models, i.e., models that do \textit{not} leverage the spatial relationships among the nodes. 
We propose a all-Multi-Layer Perceptron (all-MLP) time series forecasting architecture called \proposed{}. 
The all-MLP architecture was chosen due to its recent success in time series forecasting benchmarks. 
Furthermore, our method capitalizes on the ensemble-like behavior of deep neural networks, where each individual block within the network behaves like a base learner in an ensemble model, particularly when identity mapping residual connections are incorporated.
By integrating random projection layers into our model, we increase the diversity among the blocks' outputs, thereby improving the overall performance of the network. 
Extensive experiments conducted on the largest spatial-temporal forecasting benchmark datasets demonstrate that the proposed method outperforms alternative methods, including both spatial-temporal graph models and general forecasting models.
\end{abstract}



\maketitle

\section{Introduction}
Spatial-temporal forecasting systems are instrumental in addressing numerous real-world problems~\cite{yan2018spatial,yao2018deep,geng2019spatiotemporal,liu2020disentangling,cai2023memda}. 
Traffic flow prediction, in particular, has attracted considerable attention due to its potential to significantly impact urban planning, traffic management, and public safety~\cite{lan2022dstagnn,fan2023spatial,liu2023largest}. 
In light of this, our study specifically uses the largest traffic flow prediction benchmark dataset, LargeST~\cite{liu2023largest}, to study the spatial-temporal forecasting problem.

Conventionally, graph-based methods are utilized to solve spatial-temporal forecasting problems in the literature~\cite{liu2023largest}. 
These methods predict the future by leveraging both temporal (i.e., time series) and spatial (i.e., graph) information. 
However, it is important to note that these two types of information may not hold equal significance. 
For instance, methods like DSTAGNN~\cite{lan2022dstagnn} achieve state-of-the-art performance by relying solely on the input time series, bypassing the input graph and inferring the spatial relationship from the time series. 
As spatial relationships are learned, this type of method has the potential to uncover relationships that do not exist in the input graph.
While these methods~\cite{shang2021discrete,lan2022dstagnn,shao2022pre} have shown promising performance, they often confront computational challenges when dealing with large-scale spatial-temporal datasets. 
This is largely due to the computation of the pairwise similarity matrix (or adjacency matrix), which has quadratic space complexity relative to the number of nodes in the dataset, and forms part of the model's intermediate representation. 
This observation prompts an intriguing question: How can we maintain scalability while achieving high performance when the input graph is not utilized?

In the absence of an input graph, the spatial-temporal forecasting can be framed as a multidimensional time series forecasting problem~\cite{zeng2023transformers}. 
In this context, each node is perceived as a dimension within the multidimensional time series. 
Over the years, numerous multidimensional time series forecasting methods have been proposed~\cite{shang2021discrete,zeng2023transformers,chen2023tsmixer}. 
We are particularly interested in solutions that exclusively utilize Multi-Layer Perceptron (MLP), due to its simplicity, efficiency, and its state-of-the-art performance in multidimensional time series forecasting~\cite{chen2023tsmixer}. 
These models, often referred to as all-MLP or mixer models, comprise layers of mixer blocks~\cite{tolstikhin2021mlp,chen2023tsmixer}. 
Unlike graph-based methods, these models avoid the explicit computation of the computationally expensive pairwise similarity matrix.

Our proposed method enhances the existing mixer model's ability to capture \textit{both} the spatial and temporal aspects of the input time series. 
Specifically, we incorporate random projection into mixer models to bolster the model's capability to learn node relationships. 
It has been observed that when identity mapping connections are integrated into the model design, deep learning models exhibit ensemble-like behavior~\cite{veit2016residual}. 
This results in residual units, or mixer blocks in context of mixer models, functioning similarly to base learners in ensemble model~\cite{veit2016residual}. 
Given that diversity is a critical factor for the success of an ensemble model~\cite{margineantu1997pruning}, and diversifying the intermediate outputs is akin to diversifying the base learners in an ensemble, our proposed method has demonstrated superior performance when compared to existing solutions on large-scale spatial-temporal forecasting benchmark datasets.

\begin{figure}[htp]
\centerline{
\includegraphics[width=0.99\linewidth]{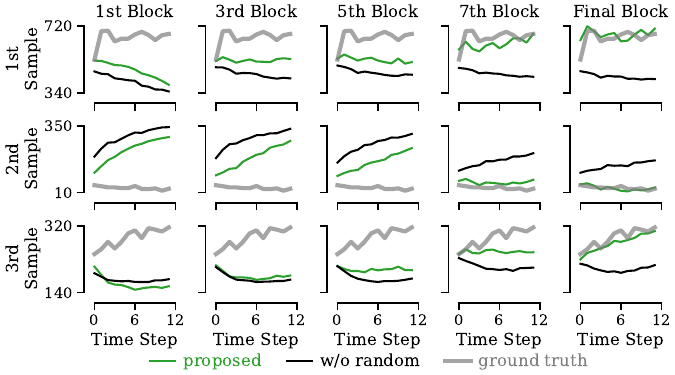}
}
\caption{
Each row consists of the intermediate and final outputs from a sampled node, while each column consists of the outputs from different mixer blocks within the model.
The three samples are from node 238, 280, and 295 of the SD dataset (see Section~\ref{sec:experiment} for details).
}
\label{fig:motivation}
\end{figure}

To demonstrate the inclusion of random projection does diversify the intermediate representation, we refer to Fig.~\ref{fig:motivation}, which presents three examples of the outputs from the first, third, fifth, seventh, and final (eighth) blocks of two mixer models. 
One of these models integrates random projection (i.e., ``proposed" in Fig.~\ref{fig:motivation}), while the other does not. 
As illustrated, the outputs of the mixer blocks in our proposed method exhibit greater diversity (i.e., the outputs from different blocks bear less similarity) compared to those of alternative models that lack random projection. 
This diversification facilitates more accurate predictions, as evidenced by the output of the final block, which closely mirrors the ground truth. 
Owing to our application of random projection in the mixer model design, we have aptly named our proposed method \proposed{}.

\begin{figure}[htp]
\centerline{
\includegraphics[width=0.99\linewidth]{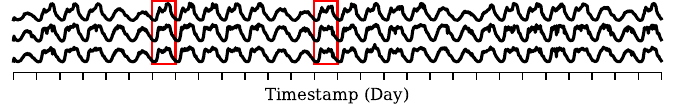}
}
\caption{
The time series for three randomly selected nodes from the LargeST dataset spans a period of four weeks.
}
\label{fig:period_example}
\end{figure}

Beyond enhancing the spatial modeling capability of mixer models, we also bolster their temporal modeling capability by processing time series in the frequency domain. 
This is particularly beneficial for spatial-temporal time series, which often exhibit periodicity, especially those influenced by human activity~\cite{yan2018spatial,yao2018deep,geng2019spatiotemporal,liu2020disentangling,cai2023memda}. 
Such periodicity is evident in traffic flow forecasting datasets, where daily and weekly patterns are common, as illustrated in Fig.~\ref{fig:period_example}. 
To effectively model these periodic signals, we employ the fast Fourier transform along with linear layers that handle complex numbers.
This approach enables us to capture patterns and trends in the data that might otherwise be overlooked by alternative designs.

The key contributions of this paper include:
\begin{itemize} 
    \item We develop a novel spatial-temporal forecasting method, \proposed{}, that does not rely on the input graph and is both effective and efficient. 
    We incorporate random projection layers in \proposed{} to enhance the spatial modeling capability of mixer models. 
    \item We enhance the temporal modeling capability of mixer models by processing time series in the frequency domain, leveraging the inherent periodicity of many spatial-temporal datasets. 
    \item Extensive experiments validate the effectiveness of our proposed method on large-scale spatial-temporal forecasting datasets, demonstrating superior performance over existing solutions. 
\end{itemize}

\section{Related Work}
In this section, we review the literature on three areas: 1) spatial-temporal forecasting methods, 2) multidimensional time series forecasting methods, and 3) random projection methods.

\sloppy
Spatial-temporal forecasting methods primarily use one of two spatial representations: 1) a graph-based format~\cite{liu2023largest}, where the spatial information is stored in a graph, and 2) a grid-based format~\cite{zhang2023mlpst}, where nodes are arranged in a grid resembling an image. 
Given that the LargeST dataset~\cite{liu2023largest} is graph-based, our review focuses on graph-based methods. 
These methods typically integrate graph modeling architectures, such as the graph convolutional network~\cite{kipf2016semi} or the graph attention network~\cite{velivckovic2017graph}, with sequential modeling architectures like recurrent neural networks~\cite{hochreiter1997long}, temporal convolutional networks~\cite{oord2016wavenet}, or transformers~\cite{vaswani2017attention}. 
Recent proposals include DCRNN~\cite{li2017diffusion}, STGCN~\cite{yu2017spatio}, ASTGCN~\cite{guo2019attention}, GWNET~\cite{wu2019graph}, AGCRN~\cite{bai2020adaptive}, STGODE~\cite{fang2021spatial}, DSTAGNN~\cite{lan2022dstagnn}, D$^2$STGNN~\cite{shao2022decoupled}, and DGCRN~\cite{li2023dynamic}. 
We incorporate all these methods into our experiments due to their relevance to our problem.
We delve into the details of these spatial-temporal forecasting methods in Appendix~\ref{apdx:baseline}.
Note, MLPST~\cite{zhang2023mlpst}, much like \proposed{}, is also a mixer-type forecasting model. 
However, it was specifically developed for grid-based spatial-temporal forecasting problems and, therefore, is not considered in this paper.

The spatial-temporal forecasting problem, in the absence of graph input, essentially transforms into a multidimensional time series forecasting problem (see Section~\ref{sec:definition} for details). 
As such, we also review methods designed for multidimensional time series forecasting. 
Over the years, transformer-based methods~\cite{vaswani2017attention} like LogTrans~\cite{li2019enhancing}, Pyraformer~\cite{liu2021pyraformer}, Autoformer~\cite{wu2021autoformer}, Informer~\cite{zhou2021informer}, and Fedformer~\cite{zhou2022fedformer} have emerged. 
However, simple linear models outperform these transformer-based methods on multidimensional long-term time series forecasting datasets as~\cite{zeng2023transformers} demonstrates. 
In contrast, mixer-based methods~\cite{tolstikhin2021mlp}, like TSMixer~\cite{chen2023tsmixer}, have demonstrated promising performance on multidimensional time series forecasting problems.
Consequently, we aim to develop a mixer-based architecture for the spatial-temporal forecasting problem.

Random projection~\cite{bingham2001random} is utilized in a variety of machine learning and data mining methods~\cite{bingham2001random,schclar2009random,cannings2017random,yeh2022embedding}.
However, the majority of these works~\cite{bingham2001random,cannings2017random,yeh2022embedding} concentrate on the efficiency gains from random projection. 
Only~\cite{schclar2009random} discusses how the diversity introduced by random projection could enhance the performance of an ensemble model, but without confirming that random projection actually introduces diversity into the model. 
Besides random projection, randomized methods like random shapelets/convolutions~\cite{renard2016east,randshape,yeh2018towards,dempster2020rocket} also perform exceptionally well for time series data on classification problems. 
Some papers also employ neural networks with partially fixed random initialized weights~\cite{huang2015trends,ulyanov2018deep,rosenfeld2019intriguing}, but these papers neither focus on the spatial-temporal forecasting problem nor provide an analysis based on the ensemble interpretation of deep neural networks~\cite{veit2016residual}. 
To the best of our knowledge, our paper is the first to investigate fixed randomized layers (or random projection) in the context of spatial-temporal forecasting problems.

\section{Definition and Problem}
\label{sec:definition}
We use lowercase letters (e.g., $x$), boldface lowercase letters (e.g., $\mathbf{x}$), uppercase letters (e.g., $X$), and boldface uppercase letters (e.g., $\mathbf{X}$) to denote scalars, vectors, matrices, and tensors, respectively. 
We begin by introducing the components of a spatial-temporal dataset. 
The spatial and temporal information are stored in the \textit{adjacency matrix} and the \textit{time series matrix}, respectively.

\begin{define} 
Given that there are $n$ entities in a spatial-temporal dataset, an adjacency matrix $A \in \mathbb{R}^{n \times n}$ stores the spatial relationships among the entities, i.e., $A[i,j]$ describes the relationship between the $i$-th entity and the $j$-th entity. 
\end{define}

\begin{define} 
Given that there are $n$ entities in a spatial-temporal dataset, a time series matrix $X \in \mathbb{R}^{n \times t}$ stores the temporal information of the entities, where $t$ is the length of the time series in the dataset. 
\end{define}

Next, we introduce the \textit{spatial-temporal forecasting} problem.

\begin{prob} 
Given an adjacency matrix $A \in \mathbb{R}^{n \times n}$ and a historical time series matrix $X_\text{past} \in \mathbb{R}^{n \times t_\text{past}}$ for the past $t_\text{past}$ steps, the goal of \textit{spatial-temporal forecasting} is to learn a model $F(\cdot)$ which predicts the future time series matrix $X_\text{future} \in \mathbb{R}^{n \times t_\text{future}}$ for the next $t_\text{future}$ steps. 
The problem can be formulated as: $F(X_\text{past}, A) \rightarrow X_\text{future}$.
\end{prob}

The problem formulation aligns with the multidimensional time series forecasting problem\footnote{Each node in the spatial-temporal data is considered a dimension of a multidimensional time series.}~\cite{zeng2023transformers} if the adjacency matrix~$A$ is disregarded by the model~$F(\cdot)$. 
It is important to note that each node in a spatial-temporal dataset can be associated with multiple feature dimensions. 
This means that the input time series to the model~$F(\cdot)$ could become a time series tensor~$\mathbf{X}_\text{past} \in \mathbb{R}^{n \times d \times t_\text{past}}$, where $d$ represents the number of features of the time series and $t_\text{past}$ is the number of time steps in the input time series.
Linear or MLP-based models can easily accommodate this additional dimensionality by reshaping the time series tensor~$\mathbf{X}_\text{past}$ into a matrix with the shape $n \times d t_\text{past}$ as they utilize fully connected (linear) layers. 
Given the $d t_\text{past}$-sized vector associated with an entity, a fully connected layer is capable of utilizing information from every dimension in each past time step for the prediction.

\section{Proposed Method}
\label{sec:method}
The proposed \proposed{} adopts an all-MLP mixer architecture akin to the TSMixer~\cite{chen2023tsmixer} method, but it differs in three aspects:

\begin{enumerate}[label={\arabic*}.]
\item Our design emphasizes the inclusion of identity mapping connections, achieved with a pre-activation design. 
The \proposed{} is inspired by the ensemble interpretation of the residual connection~\cite{veit2016residual}, which necessitates the incorporation of identity mapping connections in the design.
\item We incorporate random projection layers in our model to increase diversity among the outputs of different mixer blocks. 
This is based on the ensemble interpretation, where the mixer blocks serve a similar function to the base learners in an ensemble model. 
\item We process the time series in the frequency domain using complex linear layers, as time series generated from human activity are typically periodic. 
However, for non-periodic time series, complex layers may not be optimal. 
The determination of how periodic a signal should be to justify the use of complex linear layers is a topic for future research. 
\end{enumerate}

\begin{figure}[htp]
\centerline{
\includegraphics[width=0.6\linewidth]{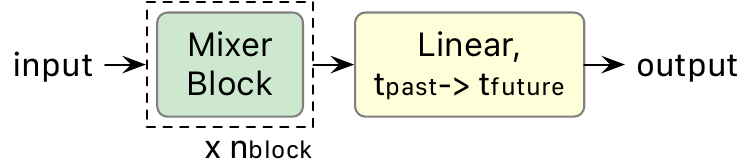}
}
\caption{
The \proposed{} architecture.
}
\label{fig:overall}
\end{figure}

The overall model design is depicted in Fig.~\ref{fig:overall}. 
Given an input historical time series matrix~$X_\text{past} \in \mathbb{R}^{n \times t_\text{past}}$, the model processes it with $n_\text{block}$ mixer blocks, which we will introduce in the next paragraph. 
The output of each mixer block is also in $\mathbb{R}^{n \times t_\text{past}}$. 
The output of the last mixer block is processed by an output linear layer to project the length of the time series to the desired length (i.e., $t_\text{future}$). 
The output time series matrix is in $\mathbb{R}^{n \times t_\text{future}}$. 
We optimize the model with mean absolute error (MAE) loss.


\begin{figure*}[t]
\centerline{
\includegraphics[width=0.75\linewidth]{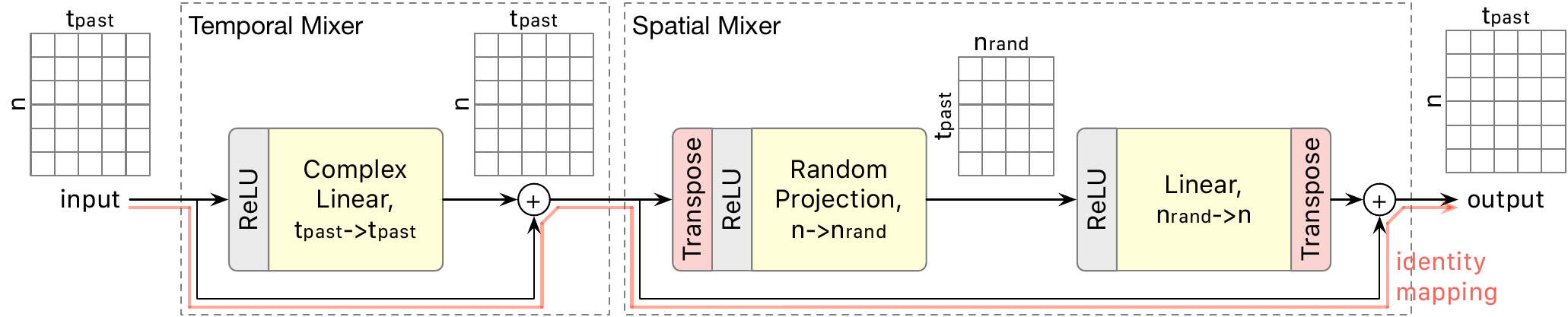}
}
\caption{
The detailed design of the mixer block.
The red line highlighted the identity mapping connection for a mixer block.
}
\label{fig:mixer}
\end{figure*}

The design of the mixer block is illustrated in Fig.~\ref{fig:mixer}. 
The highlighted forward path (in red) is used to explain the identity mapping connection in Section~\ref{sec:identity}. 
The shape of the input, intermediate, and output matrices are also included in the figure. 
The sizes of the input and output matrices are both in $\mathbb{R}^{n \times t_\text{past}}$.

The mixer block comprises two sub-blocks: 1) a temporal mixer block focusing on modeling the relationship between different time steps, and 2) a spatial mixer block focusing on modeling the relationship between different nodes. 
The temporal mixer block employs a complex linear layer after the \textsc{ReLU} activation function to model the time series in the frequency domain. 
We delve into the details of the complex linear layer in Section~\ref{sec:complex}.

The first and last operations of the spatial mixer block are matrix transpositions. 
This is done because the linear layers within the spatial mixer block are designed to model the relationship between the nodes dimension of the matrix. 
By transposing the matrix, we enable the linear layers to linearly combine the representations associated with each node, as opposed to time steps.

There are two types of linear layers in the spatial mixer block: 1) the random projection layer, and 2) the regular linear layer.
We first use the random projection layer to project the vectors within the matrix from $\mathbb{R}^n$ to $\mathbb{R}^{n_\text{rand}}$. 
We regard the random projection layer as a type of linear layer because it consists of linear operations, which we implemented with a fixed, randomly initialized linear layer. 
We consider $n_\text{rand}$ a hyper-parameter of our method and provide an empirical study of this parameter in Section~\ref{sec:parameter}.
The details about the random projection layer are discussed in Section~\ref{sec:random}. 
Subsequently, a regular linear layer is used to map the size of the vector within the matrix from $n_\text{rand}$ back to $n$. 
Both linear layers are preceded by a \textsc{ReLU} activation function.

\subsection{Complex Linear Layer}
\label{sec:complex}
\begin{figure}[htp]
\centerline{
\includegraphics[width=0.99\linewidth]{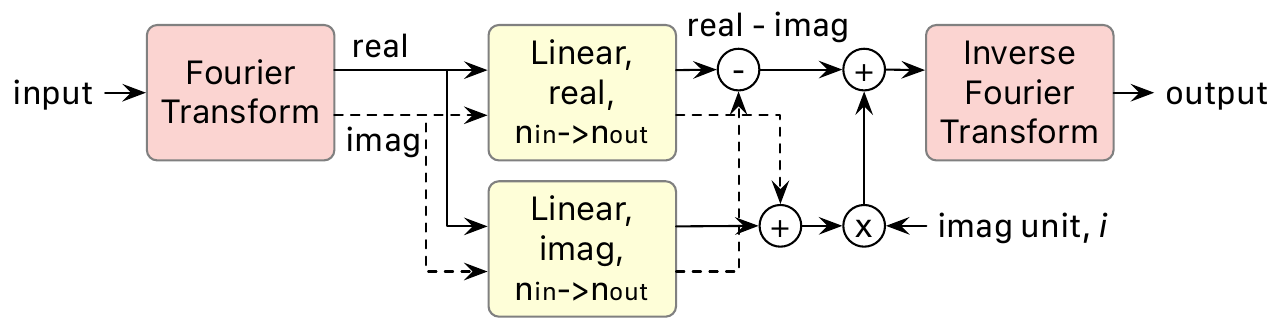}
}
\caption{
The complex linear layer.
}
\label{fig:complex}
\end{figure}

As depicted in Fig.~\ref{fig:period_example}, the time series in the large spatial-temporal forecasting benchmark dataset exhibits periodicity\footnote{Please note, periodicity is an important property that enables complex linear layers to enhance performance and ensures the predictability of a time series dataset.}. 
To leverage this periodicity, we opt to process the time series in the frequency domain using a complex linear layer~\cite{trabelsi2018deep}. 
The computational graph for the complex linear layer is illustrated in Fig.~\ref{fig:complex}. 
Initially, the input data is converted to the frequency domain using the Fast Fourier Transform (FFT) method. 
Subsequently, the real and imaginary parts are processed by two distinct linear layers, one containing the real part of the model weights and the other containing the imaginary part. 
The outputs from these two linear layers are then combined, and finally an inverse FFT layer is used to convert back to the time domain.

The design rationale stems from the fact that, unlike regular linear layers, the weights for the complex linear layer are complex numbers. 
Consider a simplified scenario where we aim to multiply input data~$x$ with a complex weight matrix $W_\text{real} + \iu W_\text{imag}$. 
As depicted in Fig.~\ref{fig:complex}, the input data~$x$ is first converted to the frequency domain as $x_\text{real} + \iu x_\text{imag}$ using the FFT method. 
Next, we multiply $x_\text{real} + \iu x_\text{imag}$ with $W_\text{real} + \iu W_\text{imag}$. 
Simple algebraic manipulation reveals that the result of this multiplication is \begin{equation} 
    (W_\text{real} x_\text{real} - W_\text{imag} x_\text{imag}) + \iu (W_\text{real} x_\text{imag} + W_\text{imag} x_\text{real})
\end{equation} 
This operation is captured in the design illustrated in Fig.~\ref{fig:complex}.

\subsection{Random Projection Layer}
\label{sec:random}
With the ensemble interpretation established in Section~\ref{sec:identity}, our goal is to utilize this interpretation to further enhance the performance of our mixer model. 
Specifically, we aim to increase the diversity among the outputs of different base learners by incorporating random projection layers.

\begin{minipage}{0.9\linewidth}
\begin{python}
import torch
import torch.nn as nn
import torch.nn.functional as F

class RPLayer(nn.Module):
    def __init__(self, in_dim, out_dim, seed):
        super(RPLayer, self).__init__()
        torch.manual_seed(seed=seed)
        weight = torch.randn(
            out_dim, in_dim, requires_grad=False)
        self.register_buffer(
            'weight', weight, persistent=True)
        self.register_buffer(
            'bias', None)

    def forward(self, x):
        return F.linear(x, self.weight, self.bias)
\end{python}
\end{minipage}

As demonstrated in the above pseudocode, the random projection layer is a fixed, randomly initialized linear layer that performs random projection~\cite{bingham2001random} on the input. 
If we split the input time series matrix~$X \in \mathbb{R}^{n \times t_\text{past}}$ into $t_\text{past}$ vectors of size $n$, the random projection layer computes $n_\text{rand}$ random combinations of the vector elements to form new vectors of size $n_\text{rand}$. 
In essence, by keeping the weights of a random projection layer fixed during training, we encourage the associated mixer block to concentrate on a random set of nodes. 
Since each random projection layer has its unique set of randomly initialized weights, different mixer blocks would focus on different sets of random nodes.
Consequently, the outputs of these mixer blocks are likely to deviate from one another.

\begin{figure}[htp]
\centerline{
\includegraphics[width=0.99\linewidth]{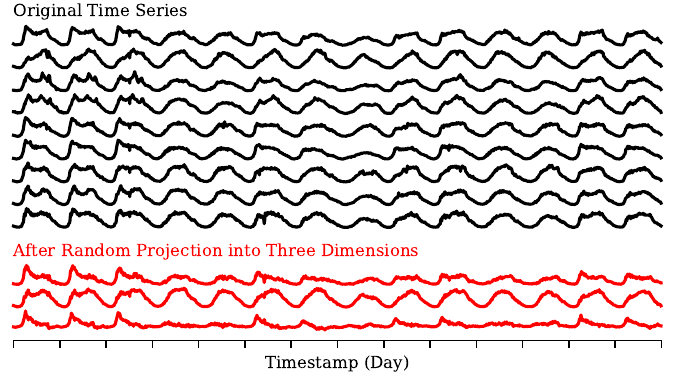}
}
\caption{
The time series output from a random projection layer retains the daily and weekly patterns of the original time series.
}
\label{fig:random_example}
\end{figure}

According to the Johnson-Lindenstrauss lemma~\cite{johnson1984extensions}, the relationships between the $t_\text{past}$ vectors are preserved after the random projection layer with high probability. 
This implies that the dynamics of the overall spatial-temporal data are largely retained in the data, even if it is projected to a much smaller space. 
To visually verify this claim, we randomly sampled nine nodes from the LargeST dataset~\cite{liu2023largest} to form a nine-dimensional time series and used random projection to reduce the dimensionality to three. 
The visualization is shown in Fig.~\ref{fig:random_example}, where the daily and weekly patterns are preserved in the time series output by random projection. 
A similar phenomenon has been observed in~\cite{yeh2023sketching} for the time series discord mining problem.

\subsection{Identity Mapping with Pre-Activation}
\label{sec:identity}
The identity mapping connection~\cite{he2016identity} plays an important role in the ensemble interpretation~\cite{veit2016residual} of the proposed \proposed{}. 
It facilitates the creation of shorter paths, which acts like base learners within the model. 
We have implemented the identity mapping connection drawing inspiration from the pre-activation design proposed in~\cite{he2016identity}, where the activation functions (e.g., \textsc{ReLU}) precede the weighted layers.

\begin{figure}[htp]
\centerline{
\includegraphics[width=0.85\linewidth]{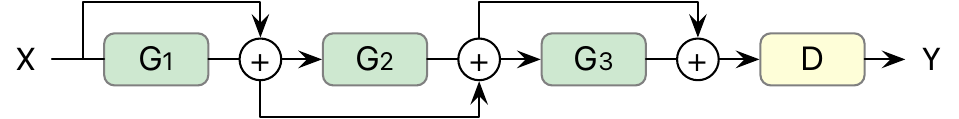}
}
\caption{
An example used for illustrating how the \proposed{} could be interpreted as an ensemble model.
In this example, there are only three mixer blocks in the model; however, the same analysis can be extended to models with more blocks.
}
\label{fig:toy}
\end{figure}

To understand why shorter paths exist within the model, we refer back to Fig.~\ref{fig:mixer}. 
We define the weighted paths for the temporal and spatial mixer sub-blocks as follows:
\begin{equation} 
    F_\text{temp}(X) \defeq \textsc{ComplexLinear}(\textsc{ReLU}(X))
\end{equation}
\begin{equation} 
    F_\text{sp}(X) \defeq \textsc{Linear}(\textsc{ReLU}(\textsc{RandProject}(\textsc{ReLU}(X^T))))^T 
\end{equation}
Given an input~$X$, the operation of the mixer block can be expressed as:
\begin{equation} 
    \textsc{Mixer}(X) = F_\text{sp}(F_\text{temp}(X) + X) + F_\text{temp}(X) + X 
\end{equation}
If we further define a function~$G(\cdot)$ as:
\begin{equation}
    G(X) \defeq F_\text{sp}(F_\text{temp}(X) + X) + F_\text{temp}(X) 
\end{equation}
The mixer block operation can be simplified to:
\begin{equation}
    \textsc{Mixer}(X) = G(X) + X 
\end{equation}
In this equation, the first term represents the weighted path and the second term is the identity mapping connection (as shown by the red line in Fig.~\ref{fig:mixer}). 
With this simplified notation in place, let us consider the case depicted in Fig.~\ref{fig:toy}, which illustrates the \proposed{} with three mixer blocks.
Here, we use $D(\cdot)$ to represent the output linear layer.
Note, the example presented in the figure only consists of three mixer blocks, but the same analysis can be extended to models with more mixer blocks.

\begin{figure}[htp]
\centerline{
\includegraphics[width=0.85\linewidth]{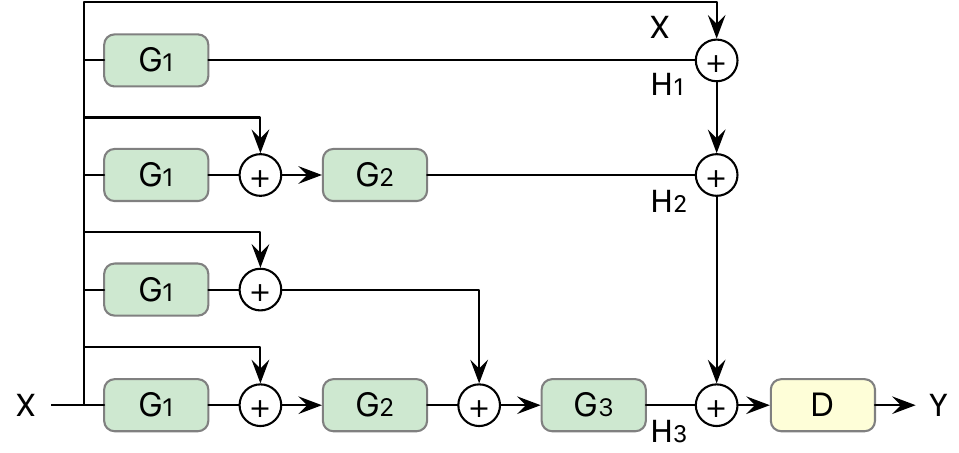}
}
\caption{
The unraveled view of different forward passes for the model illustrated in Fig.~\ref{fig:toy}.
}
\label{fig:unroll_pre}
\end{figure}

Following the analysis presented in~\cite{veit2016residual}, we can unravel the forward pass in Fig.~\ref{fig:toy} into multiple paths. 
The unraveled view is illustrated in Fig.~\ref{fig:unroll_pre}.

If we denote the outputs of $G_1(\cdot)$, $G_2(\cdot)$, and $G_3(\cdot)$ as $H_1$, $H_2$, and $H_3$ respectively, the output~$Y$ can be represented as:
$Y = D(X + H_1 + H_2 + H_3)$.
Since $D(\cdot)$ is a linear function, the above equation can be rewritten as:
$Y = D(X) + D(H_1) + D(H_2) + D(H_3)$.
If we further define: $Y_0 \defeq D(X)$, $Y_1 \defeq D(H_1)$, $Y_2 \defeq D(H_2)$, and $Y_3 \defeq D(H_3)$, it becomes clear that the prediction $Y$ is the sum of the individual predictions from each path, i.e., $Y = Y_0 + Y_1 + Y_2 + Y_3$.
In essence, the identity mapping connections introduced by the pre-activation design facilitate an ensemble-like behavior in \proposed{}.

\section{Experiment}
\label{sec:experiment}
In this section, we present experiment results that demonstrate the effectiveness of our proposed method. 
We begin by introducing the dataset, benchmark settings, and baseline methods.
Following this, we explore the benchmark results and an ablation study, which showcases the impact of each design choice.
Importantly, we show how the random projection layer significantly enhances the diversity of the network's intermediate representation, contributing substantially to the superior performance of our proposed method.
Next, we conduct a sensitivity analysis on the crucial hyper-parameters. 
It is worth noting that our proposed method is a general approach for multivariate time series forecasting. 
Thus, we also perform experiments on benchmark datasets for multivariate time series forecasting and include the results in Appendix~\ref{apdx:long_term}.
Details regarding the implementation can be found in Appendix~\ref{apdx:implementation}.
The source code for our experiments can be downloaded from~\cite{supplementary}.

\subsection{Dataset and Benchmark Setting}
\label{sec:dataset}
We conducted our experiments using the LargeST dataset~\cite{liu2023largest}, which consists of traffic data collected from 8,600 sensors in California from 2017 to 2021. 
For benchmarking purposes, we generated four sub-datasets following the procedure outlined in~\cite{liu2023largest}. 
These sub-datasets, namely SD, GBA, GLA, and CA, include sensor data from the San Diego region, the Greater Bay Area region, the Greater Los Angeles region, and all 8,600 sensors across California, respectively.
The statistics pertaining to these datasets are detailed in Appendix~\ref{apdx:dataset}.

In line with the experimental setup described in~\cite{liu2023largest}, we only utilized traffic data from 2019.
Sensor readings, originally recorded at 5-minute intervals, were aggregated into 15-minute windows, yielding 96 windows per day. 
Each sub-dataset was chronologically divided into training, validation, and test sets at a ratio of 6:2:2. 
The benchmark task was to predict the next 12 steps for each sensor at each timestamp. 
Performance was measured using the mean absolute error (MAE), root mean squared error (RMSE), and mean absolute percentage error (MAPE).

\subsection{Baseline Method}
\label{sec:baseline}
Our benchmark experiment incorporates 14 baseline methods. 
Besides the 11 baseline methods\footnote{HL, LSTM~\cite{hochreiter1997long}, DCRNN~\cite{li2017diffusion}, STGCN~\cite{yu2017spatio},  ASTGCN~\cite{guo2019attention}, GWNET~\cite{wu2019graph}, AGCRN~\cite{bai2020adaptive}, STGODE~\cite{fang2021spatial}, DSTAGNN~\cite{lan2022dstagnn}, D$^2$STGNN~\cite{shao2022decoupled}, and DGCRN~\cite{li2023dynamic}} evaluated by Liu et al.~\cite{liu2023largest}, we also explore general time series forecasting baseline methods such as the one-nearest-neighbor regressor (1NN), linear model, and TSMixer~\cite{chen2023tsmixer}. 
It is important to note that certain baseline methods are not included for the CA dataset due to scalability issues preventing the completion of these methods' experiments. 
The details about these methods are included in Appendix~\ref{apdx:baseline}.

\begin{table*}[pt]
\caption{Performance comparisons. 
We bold the best-performing results.
The performance reported in the ``Average" column is computed by averaging over 12 predicted time steps.
The absence of baselines on the GLA and CA datasets indicates that the models incur an out-of-memory issue.
Param: the number of learnable parameters. K: $10^3$. M: $10^6$.}
\label{tab:largest_benchmark}
\begin{center}
\resizebox*{!}{0.93\textheight}{%
\begin{tabular}{llcccc|ccc|ccc|ccc}
    \shline
    \multirow{2}{*}{Data} & \multirow{2}{*}{Method} & \multirow{2}{*}{Param} & \multicolumn{3}{c}{Horizon 3} & \multicolumn{3}{c}{Horizon 6} & \multicolumn{3}{c}{Horizon 12} & \multicolumn{3}{c}{Average} \\ \cline{4-15} 
     &  &  & MAE & RMSE & MAPE & MAE & RMSE & MAPE & MAE & RMSE & MAPE & MAE & RMSE & MAPE \\ 
     \hline \hline
     \multirow{15}{*}{SD} & HL & -- & 33.61 & 50.97 & 20.77\% & 57.80 & 84.92 & 37.73\% & 101.74 & 140.14 & 76.84\% & 60.79 & 87.40 & 41.88\% \\
     & LSTM & 98K & 19.17 & 30.75 & 11.85\% & 26.11 & 41.28 & 16.53\% & 38.06 & 59.63 & 25.07\% & 26.73 & 42.14 & 17.17\% \\
     & ASTGCN & 2.2M & 20.09 & 32.13 & 13.61\% & 25.58 & 40.41 & 17.44\% & 32.86 & 52.05 & 26.00\% & 25.10 & 39.91 & 18.05\% \\
     & DCRNN & 373K & 17.01 & 27.33 & 10.96\% & 20.80 & 33.03 & 13.72\% & 26.77 & 42.49 & 18.57\% & 20.86 & 33.13 & 13.94\% \\
     & AGCRN & 761K & 16.05 & 28.78 & 11.74\% & 18.37 & 32.44 & 13.37\% & 22.12 & 40.37 & 16.63\% & 18.43 & 32.97 & 13.51\% \\
     & STGCN & 508K & 18.23 & 30.60 & 13.75\% & 20.34 & 34.42 & 15.10\% & 23.56 & 41.70 & 17.08\% & 20.35 & 34.70 & 15.13\% \\
     & GWNET & 311K & 15.49 & 25.45 & \textbf{9.90\%} & 18.17 & 30.16 & 11.98\% & 22.18 & 37.82 & 15.41\% & 18.12 & 30.21 & 12.08\% \\
     & STGODE & 729K & 16.76 & 27.26 & 10.95\% & 19.79 & 32.91 & 13.18\% & 23.60 & 41.32 & 16.60\% & 19.52 & 32.76 & 13.22\% \\
     & DSTAGNN & 3.9M & 17.83 & 28.60 & 11.08\% & 21.95 & 35.37 & 14.55\% & 26.83 & 46.39 & 19.62\% & 21.52 & 35.67 & 14.52\% \\
     & DGCRN & 243K & 15.24 & 25.46 & 10.09\% & 17.66 & 29.65 & 11.77\% & 21.38 & 36.67 & 14.75\% & 17.65 & 29.70 & 11.89\% \\
     & D$^2$STGNN & 406K & \textbf{14.85} & 24.95 & 9.91\% & 17.28 & 29.05 & 12.17\% & 21.59 & 35.55 & 16.88\% & 17.38 & 28.92 & 12.43\% \\
     \cline{2-15}
     & 1NN & - & 21.79 & 35.15 & 13.79\% & 25.64 & 41.59 & 17.05\% & 30.77 & 50.59 & 22.38\% & 25.47 & 41.36 & 17.25\% \\
     & Linear & 3.5K & 20.58 & 33.30 & 12.98\% & 27.00 & 43.87 & 18.20\% & 32.35 & 53.51 & 22.38\% & 25.85 & 42.35 & 17.10\% \\
     & TSMixer & 815K & 17.13 & 27.42 & 11.35\% & 19.30 & 31.07 & 12.50\% & 22.03 & 35.70 & 14.26\% & 19.06 & 30.66 & 12.55\% \\
     \cline{2-15}
     & \proposed{} & 1.5M & 15.12 & \textbf{24.83} & 9.97\% & \textbf{17.04} & \textbf{28.24} & \textbf{10.98\%} & \textbf{19.60} & \textbf{32.96} & \textbf{13.12\%} & \textbf{16.90} & \textbf{27.97} & \textbf{11.07\%} \\
     \hline \hline
     \multirow{15}{*}{GBA} & HL & -- & 32.57 & 48.42 & 22.78\% & 53.79 & 77.08 & 43.01\% & 92.64 & 126.22 & 92.85\% & 56.44 & 79.82 & 48.87\% \\
     & LSTM & 98K & 20.41 & 33.47 & 15.60\% & 27.50 & 43.64 & 23.25\% & 38.85 & 60.46 & 37.47\% & 27.88 & 44.23 & 24.31\% \\
     & ASTGCN & 22.3M & 21.40 & 33.61 & 17.65\% & 26.70 & 40.75 & 24.02\% & 33.64 & 51.21 & 31.15\% & 26.15 & 40.25 & 23.29\% \\
     & DCRNN & 373K & 18.25 & 29.73 & 14.37\% & 22.25 & 35.04 & 19.82\% & 28.68 & 44.39 & 28.69\% & 22.35 & 35.26 & 20.15\% \\
     & AGCRN & 777K & 18.11 & 30.19 & 13.64\% & 20.86 & 34.42 & 16.24\% & 24.06 & 39.47 & 19.29\% & 20.55 & 33.91 & 16.06\% \\
     & STGCN & 1.3M & 20.62 & 33.81 & 15.84\% & 23.19 & 37.96 & 18.09\% & 26.53 & 43.88 & 21.77\% & 23.03 & 37.82 & 18.20\% \\
     & GWNET & 344K & 17.74 & 28.92 & 14.37\% & 20.98 & 33.50 & 17.77\% & 25.39 & 40.30 & 22.99\% & 20.78 & 33.32 & 17.76\% \\
     & STGODE & 788K & 18.80 & 30.53 & 15.67\% & 22.19 & 35.91 & 18.54\% & 26.27 & 43.07 & 22.71\% & 21.86 & 35.57 & 18.33\% \\
     & DSTAGNN & 26.9M & 19.87 & 31.54 & 16.85\% & 23.89 & 38.11 & 19.53\% & 28.48 & 44.65 & 24.65\% & 23.39 & 37.07 & 19.58\% \\
     & DGCRN & 374K & 18.09 & 29.27 & 15.32\% & 21.18 & 33.78 & 18.59\% & 25.73 & 40.88 & 23.67\% & 21.10 & 33.76 & 18.58\% \\
     & D$^2$STGNN & 446K & \textbf{17.20} & \textbf{28.50} & \textbf{12.22\%} & 20.80 & 33.53 & \textbf{15.32\%} & 25.72 & 40.90 & 19.90\% & 20.71 & 33.44 & 15.23\% \\
     \cline{2-15}
     & 1NN & - & 24.84 & 41.30 & 17.70\% & 29.31 & 48.56 & 22.92\% & 35.22 & 58.44 & 31.07\% & 29.10 & 48.23 & 23.14\% \\
     & Linear & 3.5K & 21.55 & 34.79 & 17.94\% & 27.24 & 43.36 & 23.66\% & 31.50 & 51.56 & 26.18\% & 26.12 & 42.14 & 22.10\% \\
     & TSMixer & 3.1M & 17.57 & 29.22 & 14.14\% & 19.85 & 32.64 & 16.95\% & 22.27 & 37.60 & 18.63\% & 19.58 & 32.56 & 16.58\% \\
     \cline{2-15}
     & \proposed{} & 2.3M & 17.35 & 28.69 & 13.42\% & \textbf{19.44} & \textbf{32.04} & 15.61\% & \textbf{21.65} & \textbf{36.20} & \textbf{17.42\%} & \textbf{19.06} & \textbf{31.54} & \textbf{15.09\%} \\
     \hline \hline
     \multirow{13}{*}{GLA} & HL & -- & 33.66 & 50.91 & 19.16\% & 56.88 & 83.54 & 34.85\% & 98.45 & 137.52 & 71.14\% & 59.58 & 86.19 & 38.76\% \\
     & LSTM & 98K & 20.09 & 32.41 & 11.82\% & 27.80 & 44.10 & 16.52\% & 39.61 & 61.57 & 25.63\% & 28.12 & 44.40 & 17.31\% \\
     & ASTGCN & 59.1M & 21.11 & 34.04 & 12.29\% & 28.65 & 44.67 & 17.79\% & 39.39 & 59.31 & 28.03\% & 28.44 & 44.13 & 18.62\% \\
     & DCRNN & 373K & 18.33 & 29.13 & 10.78\% & 22.70 & 35.55 & 13.74\% & 29.45 & 45.88 & 18.87\% & 22.73 & 35.65 & 13.97\% \\
     & AGCRN & 792K & 17.57 & 30.83 & 10.86\% & 20.79 & 36.09 & 13.11\% & 25.01 & 44.82 & 16.11\% & 20.61 & 36.23 & 12.99\% \\
     & STGCN & 2.1M & 19.87 & 34.01 & 12.58\% & 22.54 & 38.57 & 13.94\% & 26.48 & 45.61 & 16.92\% & 22.48 & 38.55 & 14.15\% \\
     & GWNET & 374K & 17.30 & 27.72 & 10.69\% & 21.22 & 33.64 & 13.48\% & 27.25 & 43.03 & 18.49\% & 21.23 & 33.68 & 13.72\% \\
     & STGODE & 841K & 18.46 & 30.05 & 11.94\% & 22.24 & 36.68 & 14.67\% & 27.14 & 45.38 & 19.12\% & 22.02 & 36.34 & 14.93\% \\
     & DSTAGNN & 66.3M & 19.35 & 30.55 & 11.33\% & 24.22 & 38.19 & 15.90\% & 30.32 & 48.37 & 23.51\% & 23.87 & 37.88 & 15.36\% \\
     \cline{2-15}
     & 1NN & - & 23.23 & 38.69 & 13.44\% & 27.75 & 45.92 & 17.07\% & 33.49 & 55.51 & 22.86\% & 27.49 & 45.57 & 17.28\% \\
     & Linear & 3.5K & 21.32 & 34.48 & 13.35\% & 27.45 & 43.83 & 17.79\% & 32.50 & 52.69 & 21.76\% & 26.40 & 42.56 & 17.16\% \\
     & TSMixer & 4.6M & 20.38 & 224.82 & 13.62\% & 22.90 & 229.86 & 15.51\% & 23.63 & 135.09 & 15.56\% & 22.12 & 207.68 & 14.87\% \\
     \cline{2-15}
     & \proposed{} & 3.2M & \textbf{16.49} & \textbf{26.75} & \textbf{9.75\%} & \textbf{18.82} & \textbf{30.56} & \textbf{11.58\%} & \textbf{21.18} & \textbf{35.10} & \textbf{13.46\%} & \textbf{18.46} & \textbf{30.13} & \textbf{11.34\%} \\
     \hline \hline
     \multirow{10}{*}{CA} & HL & -- & 30.72 & 46.96 & 20.43\% & 51.56 & 76.48 & 37.22\% & 89.31 & 125.71 & 76.80\% & 54.10 & 78.97 & 41.61\% \\
     & LSTM & 98K & 19.01 & 31.21 & 13.57\% & 26.49 & 42.54 & 20.62\% & 38.41 & 60.42 & 31.03\% & 26.95 & 43.07 & 21.18\% \\
     & DCRNN & 373K & 17.52 & 28.18 & 12.55\% & 21.72 & 34.19 & 16.56\% & 28.45 & 44.23 & 23.57\% & 21.81 & 34.35 & 16.92\% \\
     & STGCN & 4.5M & 19.14 & 32.64 & 14.23\% & 21.65 & 36.94 & 16.09\% & 24.86 & 42.61 & 19.14\% & 21.48 & 36.69 & 16.16\% \\
     & GWNET & 469K & 16.93 & 27.53 & 13.14\% & 21.08 & 33.52 & 16.73\% & 27.37 & 42.65 & 22.50\% & 21.08 & 33.43 & 16.86\% \\
     & STGODE & 1.0M & 17.59 & 31.04 & 13.28\% & 20.92 & 36.65 & 16.23\% & 25.34 & 45.10 & 20.56\% & 20.72 & 36.65 & 16.19\% \\
     \cline{2-15}
     & 1NN & - & 21.88 & 36.67 & 15.19\% & 25.94 & 43.36 & 19.19\% & 31.29 & 52.52 & 25.65\% & 25.76 & 43.10 & 19.43\% \\
     & Linear & 3.5K & 19.82 & 32.39 & 14.73\% & 25.20 & 40.97 & 19.24\% & 29.80 & 49.34 & 23.09\% & 24.32 & 39.88 & 18.52\% \\
     & TSMixer & 9.5M & 18.40 & 106.28 & 14.30\% & 19.77 & 73.98 & 15.30\% & 22.56 & 87.56 & 17.80\% & 19.86 & 90.20 & 15.79\% \\
     \cline{2-15}
     & \proposed{} & 7.8M & \textbf{15.90} & \textbf{26.08} & \textbf{11.69\%} & \textbf{17.79} & \textbf{29.37} & \textbf{13.23\%} & \textbf{19.93} & \textbf{33.18} & \textbf{15.11\%} & \textbf{17.50} & \textbf{28.90} & \textbf{13.03\%} \\
     \shline
\end{tabular}%
}
\end{center}
\end{table*}

\subsection{Benchmark Result}
The benchmark results are summarized in Table~\ref{tab:largest_benchmark}. 
Our discussion on the performance differences among various methods will proceed in three parts: firstly, we will discuss the 11 baselines included in the original LargeST benchmark~\cite{liu2023largest}; next, we will address the three new time series forecasting baselines we added to the benchmark; finally, we will highlight the performance of our proposed \proposed{} method in comparison to the baselines.

Regarding the original 11 baseline methods, HL and LSTM are typically outperformed by other methods as they do not consider inter-node relationships like the remaining nine spatial-temporal methods. 
Newer methods such as DGCRN and D$^2$STGNN usually outperform other baselines. 
However, these two methods cannot be scaled to the two larger datasets, GLA and CA. 
Among older methods, AGCRN and GWNET outperform others on the SD, GBA, and GLA datasets. 
When considering the CA dataset, only four spatial-temporal models are applicable; GWNET, STGCN, and STGODE performance are comparable with each other.

In the case of the three new baselines (1NN, Linear, and TSMixer), both 1NN and Linear perform poorly for the same reason as the HL and LSTM baselines: these methods do not model the relationship between the nodes. 
When comparing TSMixer with the original 11 baseline methods, it is comparable with the best method on the two smaller datasets (SD and GBA). 
However, it performs noticeably worse in terms of RMSE on the two larger datasets (GLA and CA). 
A possible reason is that TSMixer is overfitting to MAE, which also serves as the loss function when training the model. 
Despite TSMixer's high parameter count, it can still be applied to the largest dataset (CA) because the memory scales linearly with respect to the number of nodes, unlike the more expensive graph-based baselines (ASTGCN, AGCRN, DSTAGNN, DGCRN, and D$^2$STGNN).

For smaller datasets such as SD and GBA, \proposed{} surpasses the performance of baseline methods in the later time steps of the predictions. 
When it comes to larger datasets like GLA and CA, \proposed{} outperforms all baseline methods across every performance measure and time horizon. 
Even though the proposed method has a high number of parameters similar to TSMixer, it is not overfitting to the loss function and performs exceptionally well across the board. 
The memory complexity of \proposed{} scales linearly with respect to the number of nodes, enabling it to be applied to the largest dataset (CA).

\subsection{Ablation Study}
\label{sec:ablation}
\begin{figure}[htp]
\centerline{
\includegraphics[width=0.99\linewidth]{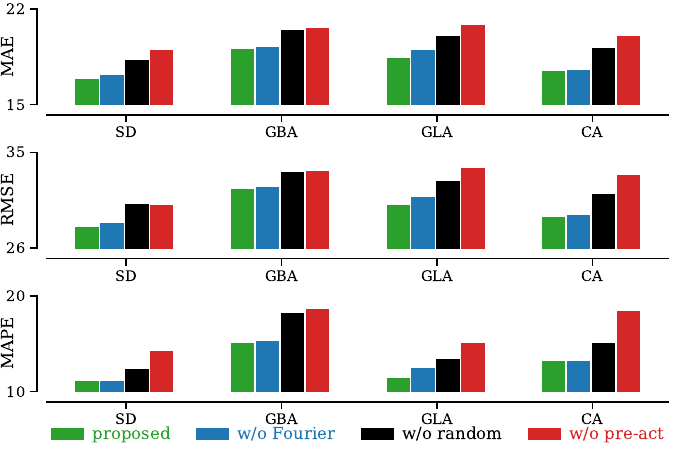}
}
\caption{
The ablation study result measured using average MAE, RMSE, and MAPE.
Removing each component resulted in a degradation of the performance.
}
\label{fig:ablation_avg}
\end{figure}

We conducted an ablation study on the three most crucial design decisions associated with our proposed \proposed{} method. 
The three decisions were: 1) implementing the identity mapping connection with pre-activation, 2) introducing random projection layers, and 3) processing the time series in the frequency domain.
The results of the ablation study are presented in Fig.~\ref{fig:ablation_avg}.

The pre-activation design emerged as the most consequential design decision. 
Upon its replacement with a post-activation design, we effectively removed the most important element of the original design, leading to a significant performance decrease. 
This decline is likely due to the elimination of identity mapping connections from the network.
These connections are vital for the ensemble interpretation of the residual connection. 
Without them, the inclusion of a random projection layer no longer makes sense, as it weakens the ensemble-like behavior of the residual blocks (i.e., the mixer blocks in \proposed{}), which was the motivation behind the design of the random projection layer.

The random projection layer design was the second most important decision in terms of performance difference. 
We disabled the random projection layer by converting it into a regular linear layer with trainable weights. 
This layer helps each mixer block in focusing on different aspects of the inter-node relationship, so its importance is not surprising. 
To verify our claim that the random projection layer promotes diversity, we constructed a correlation-error diagram comparing the proposed method with the variant without random projection layer. 
This diagram serves a similar purpose to the kappa-error diagram~\cite{margineantu1997pruning} used for analyzing base learners in an ensemble model. 
Before discussing the diagram, let us first introduce what a correlation-error diagram is.

A correlation-error diagram is a visualization tool akin to the kappa-error diagram~\cite{margineantu1997pruning}, used for analyzing the trade-off between performance and diversity in ensemble models. 
Each dot in the diagram represents a pair of base learners, with the $y$-axis showing the average performance of the pair's outputs and the $x$-axis indicating the degree of agreement between them. 
Specifically, we use the Pearson correlation coefficient to measure agreement between the outputs of a pair of mixer blocks (i.e., the base learners). 
We use MAE, RMSE, and MAPE as performance measures. 
The figure created using MAE and Pearson correlation coefficient is shown in Fig.~\ref{fig:corr_error_mae}. 
Figures using other performance measures are included in Appendix~\ref{apdx:ablation}. 
We observe that the dots associated with the proposed method with a random projection layer have a wider distribution (larger range in $y$-axis) compared to the variant without the random projection layer.
This confirms that our random projection layer effectively increases the diversity of the intermediate representation.

\begin{figure}[htp]
\centerline{
\includegraphics[width=0.99\linewidth]{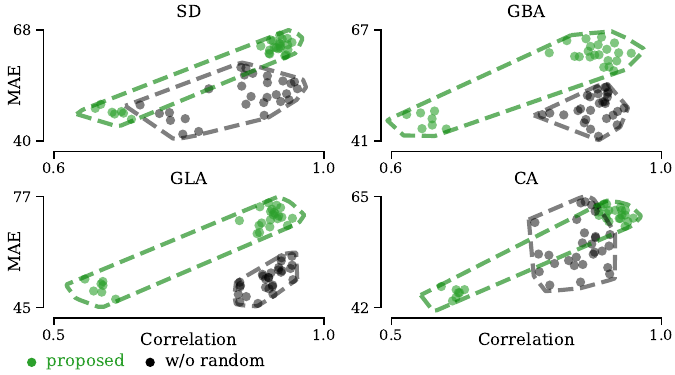}
}
\caption{
The correlation-error diagram illustrates the interplay of diversity and error (MAE) across individual blocks in the model.
}
\label{fig:corr_error_mae}
\end{figure}

The decision to process the time series in the frequency domain had the least impact on performance. 
The performance improvement is minor compared to the other two design choices. 
One possible reason is that the Fourier transformation is a linear transformation, so the majority of its benefits could be learned by the model during training.
We also analyze the effect of different design choices at various time horizons and the findings are reported in Appendix~\ref{apdx:ablation}.

\subsection{Parameter Sensitivity Analysis}
\label{sec:parameter}
\begin{figure}[htp]
\centerline{
\includegraphics[width=0.99\linewidth]{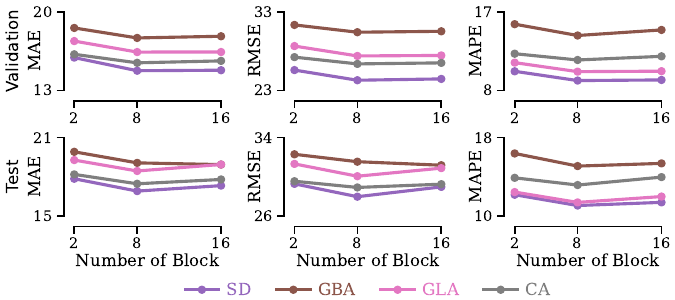}
}
\caption{
The parameter sensitivity analysis for the \textit{number of block} parameter measured using average MAE, RMSE, and MAPE.
Setting the number of block to eight generally yield better result.
This observation holds true for both validation and test data.
}
\label{fig:param_layer_avg}
\end{figure}

In our parameter sensitivity analysis, we focus on two hyper-parameters: the number of mixer blocks and the number of neurons in the random projection layer. 
We report performance on both the validation and test sets to demonstrate the generalizability of our findings across different data partitions. 
Fig.~\ref{fig:param_layer_avg} illustrates the parameter sensitivity analysis for the number of mixer blocks.

Setting the number of blocks to eight generally yields the best performance on both the validation and test data. 
Reducing the number of blocks to two significantly diminishes performance, indicating the necessity of multiple blocks to achieve ensemble-like behavior. 
Increasing the number of blocks to 16 results in minor performance improvement on some datasets. 
However, as it doubles the model size, the minor gain in performance may not justify the increased computational cost. 
Similar trends are observed on both the validation and test data, suggesting that we can use the validation set to tune this hyper-parameter.

\begin{figure}[htp]
\centerline{
\includegraphics[width=0.99\linewidth]{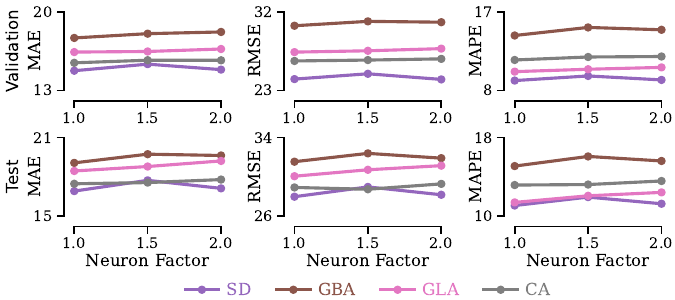}
}
\caption{
The parameter sensitivity analysis for the \textit{number of neurons} parameter measured using average MAE, RMSE, and MAPE.
The number is set by $m_{neuron}\sqrt{n}$, where $n$ is the number of nodes in the graph and $m_{neuron}$ is a factor. 
Setting this factor to one often has the best performance.
This observation holds true for both validation and test data.
}
\label{fig:param_neuron_avg}
\end{figure}

For the number of neurons in the random projection layer, we set it as a function with respect to the number of nodes in the spatial-temporal data, as more neurons may be required for datasets with larger number of nodes. 
If the graph has $n$ nodes, we set the number of neurons in the random projection layer as $m_{neuron} \sqrt{n}$, where $m_{neuron}$ is the hyper-parameter controlling the number of neurons in the layer. 
This design allows the model to automatically use more neurons for datasets with more number of nodes, even if the hyper-parameter $m_{neuron}$ is set to the same value for all datasets. 
Fig.~\ref{fig:param_neuron_avg} displays the results of the parameter sensitivity analysis for the hyper-parameter $m_{neuron}$.

We evaluated the model under three different settings of $m_{neuron}$: $\{1.0, 1.5, 2.0\}$.
Our observation indicates that setting the hyper-parameter to 1.0 generally leads to good performance, although for some datasets, a setting of 2.0 may yield better results. 
However, the performance difference is minimal and may not justify the additional computational cost associated with the 2.0 setting. 
Additionally, the trends on the validation data closely mirror those on the testing data, justifying the use of validation data for setting the hyper-parameter.

\section{Conclusion}
In this paper, we proposed \proposed{}, an all-MLP mixer model that incorporates random projection layers. 
These layers enhance the diversity among the outputs of each mixer block, thereby improving the overall performance of the model. 
Our experiments demonstrated that the random projection layers not only improve the diversity of the intermediate representation but also boost the model's overall performance on large-scale spatial-temporal benchmark datasets in the literature~\cite{liu2023largest}. 
In future work, we plan to investigate the potential of applying time series foundation models~\cite{yeh2023toward} for tackling spatial-temporal forecasting problems.

\bibliographystyle{ACM-Reference-Format}
\bibliography{section/reference}

\clearpage
\appendix

\section{Supplementary}
In this section, we provide supplementary materials for the paper. 
These materials encompass additional details and results that were excluded from the primary text due to space constraints.

\subsection{Dataset Statistics}
\label{apdx:dataset}
The statistics for the SD, GBA, GLA, and CA dataset are summarized in Table~\ref{tab:dataset}.

\begin{table}[htp]
\caption{Dataset statistics.}
\label{tab:dataset}
\begin{center}
\begin{tabular}{lccc}
\shline
Data & \# of nodes & \# of time steps & time range \\ \hline 
SD & 716 & 35,040 & [1/1/2019, 1/1/2020) \\
GBA & 2,352 & 35,040 & [1/1/2019, 1/1/2020) \\
GLA & 3,834 & 35,040 & [1/1/2019, 1/1/2020) \\
CA & 8,600 & 35,040 & [1/1/2019, 1/1/2020) \\ \shline
\end{tabular}
\end{center}
\end{table} 

\subsection{Baseline Method}
\label{apdx:baseline}
The details about the 14 baseline methods are provided below:

\begin{itemize}
\item \textbf{HL}: 
The prediction for all future time steps is generated by using the last value from the historical data.
\item \textbf{LSTM}~\cite{hochreiter1997long}:
The Long Short-Term Memory (LSTM), a variant of the recurrent neural network architecture, is specifically designed to process sequential data, making it a general method for time series forecasting. 
When deploying this architecture on the LargeST dataset, the same model weights are utilized across the time series of different nodes.
\item \textbf{ASTGCN}~\cite{guo2019attention}: 
The Attention-based Spatial-Temporal Graph Convolutional Network (ASTGCN) model employs a spatial-temporal attention mechanism to capture spatial-temporal correlations. 
It also uses graph convolutions and conventional convolutions to extract spatial and temporal patterns, respectively.
\item \textbf{DCRNN}~\cite{li2017diffusion}:
\sloppy
The Diffusion Convolutional Recurrent Neural Network (DCRNN) incorporates diffusion convolution, the sequence-to-sequence architecture, and the scheduled sampling technique to capture spatial-temporal patterns.
\item \textbf{AGCRN}~\cite{bai2020adaptive}:
The Adaptive Graph Convolutional Recurrent Network (AGCRN) extends the design of recurrent networks with a node adaptive parameter learning module, enabling the capture of node-specific patterns. 
Additionally, it includes a data adaptive graph generation module to infer the inter-dependencies among different nodes.
\item \textbf{STGCN}~\cite{yu2017spatio}:
The Spatial-Temporal Graph Convolutional Networks (STGCN) employs both graph and temporal convolutions to model the spatial-temporal correlations within the data.
\item \textbf{GWNET}~\cite{wu2019graph}: 
The Graph WaveNet (GWNET) is another network that relies on convolution operations. 
Specifically, it employs a gated temporal convolution module and a graph convolution layer to model the spatial-temporal correlation.
\item \textbf{STGODE}~\cite{fang2021spatial}:
The Spatial-Temporal Graph Ordinary Differential Equation Networks (STGODE) models spatial-temporal dynamics using tensor-based ordinary differential equations. 
The model design also incorporates a semantical adjacency matrix and temporal dilated convolution modules.
\item \textbf{DSTAGNN}~\cite{lan2022dstagnn}:
The Dynamic Spatial-Temporal Aware Graph Neural Network (DSTAGNN) is a method that does not use a predefined static adjacency matrix. 
Instead, it learns the dynamic spatial associations among nodes and utilizes a spatial-temporal attention module based on multi-order Chebyshev polynomials to capture these associations. 
To model the temporal associations, it employs gated convolution modules.
\item \textbf{DGCRN}~\cite{li2023dynamic}:
The Dynamic Graph Convolutional Recurrent Network (DGCRN) model combines graph convolution networks with recurrent networks. 
In this model, a dynamic adjacency matrix is progressively generated by a hyper-network in synchronization with the recurrent steps. 
This dynamic adjacency matrix, in conjunction with the predefined static adjacency matrix, is utilized to generate predictions.
\item \textbf{D$^2$STGNN}~\cite{shao2022decoupled}:
The Decoupled Dynamic Spatial-Temporal Graph Neural Network (D$^2$STGNN) decouples the diffusion and inherent information in a data-driven manner using an estimation gate and a residual decomposition mechanism. 
Additionally, it employs a dynamic graph learning module that learns the dynamic characteristics of the spatial-temporal graph.
\item \textbf{1NN}: 
The 1-Nearest-Neighbor (1NN) method serves as a simple baseline for a range of time series problems~\cite{bagnall2017great,dau2019ucr}. 
We have implemented this baseline by leveraging the matrix profile for its efficiency~\cite{yeh2016matrix,zimmerman2019matrix}. 
Notably, the version of the 1NN method benchmarked in this study is the most basic form, where each node is treated as independent from the others. 
While more advanced techniques from the literature, such as those presented in~\cite{yeh2017matrix,martinez2019methodology,yeh2022error}, could potentially enhance the method, we plan to explore these techniques in our future work.
\item \textbf{Linear}:
Having proven its effectiveness in general time series forecasting~\cite{zeng2023transformers}, the linear model has been incorporated into our benchmark experiments. 
Similar to LSTM, the same model weights are used across the time series of various nodes.
\item \textbf{TSMixer}~\cite{chen2023tsmixer}: 
The Time Series Mixer (TSMixer) is a stacked Multi-Layer Perceptron (MLP) that efficiently extracts information by utilizing mixing operations across both the time and feature dimensions (i.e., nodes for spatial-temporal data). 
These mixing operations are capable of capturing the relationships between different time steps and nodes.
\end{itemize}

\subsection{Implementation Detail}
\label{apdx:implementation}
The experiments were carried out using Python 3.10.11 on a Linux server equipped with an AMD EPYC 7713 64-Core Processor and NVIDIA Tesla A100 GPU. 
Our implementation leverages PyTorch 2.0.1 to realize the proposed \proposed{}, TSMixer, and the Linear model.
All three models are trained using the AdamW optimizer~\cite{loshchilov2017decoupled} with default hyper-parameter settings. 
The MAE loss function was used, following~\cite{liu2023largest}.
We have incorporated an early stopping mechanism, with the patience parameter set to seven.
In the case of the proposed method, we configure the number of mixer blocks to eight and set the random projection dimension to $\sqrt{n}$, where $n$ is the number of nodes in the graph.
For TSMixer, the number of mixer blocks is also set to eight, and the number of hidden dimensions is fixed at 64, in accordance with the parameterization used by the original author in the Traffic prediction dataset experiment~\cite{chen2023tsmixer}.
The 1NN method is implemented using pyscamp 0.4.0~\cite{zimmerman2019matrix}. 
1NN was not tuned because there are no hyper-parameters in our implementation. 
Further details on our implementation can be found in the released source code~\cite{supplementary}.
The results of LSTM~\cite{hochreiter1997long}, ASTGCN~\cite{guo2019attention}, AGCRN~\cite{bai2020adaptive}, DSTAGNN~\cite{lan2022dstagnn}, DGCRN~\cite{li2023dynamic}, and D$^2$STGNN~\cite{shao2022decoupled} were obtained from~\cite{liu2023largest}.

\subsection{Additional Discussion about Table~\ref{tab:largest_benchmark}}
In this section, we delve into a more comprehensive discussion about the benchmark results shown in Table~\ref{tab:largest_benchmark}.

Fist, we examine the different graph neural network-based methods.
This analysis is performed using the average performances. 
The best graph method on the SD dataset is D$^2$STGNN, which outperforms the others on both MAE and RMSE. 
For MAPE, DGCRN outperforms the others. 
Both D$^2$STGNN and DGCRN have an adjacency matrix learning component, suggesting that modeling the cross-variate dependency is important for the forecasting problem. 
For the GBA dataset, AGCRN, GWNET, and D$^2$STGNN achieve the best MAE, RMSE, and MAPE, respectively. 
These methods once again consist of an adjacency matrix learning component. 
For the GLA dataset, AGCRN is the best graph method, achieving the best MAE and MAPE, while GWNET achieves the best RMSE. 
Both methods are capable of adjacency matrix learning. 
For the CA dataset, STGODE, DCRNN, and STGCN achieve the best MAE, RMSE, and MAPE, respectively. 
These three methods outperform the others because the stronger graph alternatives are not applicable for the CA dataset due to scalability issues.

Next, we compared the proposed \proposed{} with the second-best results on different datasets.
According to the average MAE, RMSE, and MAPE, the second-best results on the SD dataset are from D$^2$STGNN and DGCRN, the second-best results on the GBA dataset are from D$^2$STGNN and TSMixer, the second-best results on the GLA dataset are from AGCRN and GWNET, and the second-best results on the CA dataset are from GWNET and TSMixer. 
We are providing a detailed analysis of the similarities and differences between our proposed method and AGCRN, D$^2$STGNN, DGCRN, GWNET, and TSMixer in the following.

AGCRN is a graph convolutional recurrent network with two special designs: 1) node adaptive parameter learning and 2) data adaptive graph generation. 
The major difference between AGCRN and our method is that AGCRN uses graph convolution and recurrent operations to process the input signal, while our method uses only linear layers to process the signal. 
AGCRN utilizes data adaptive graph generation techniques to learn the dependency between different nodes, while our method learns such information with spatial mixers. 
However, for both methods, the ability to learn inter-node dependency is crucial for achieving competitive performances. 
Note that the data adaptive graph generation module needs to compute an $n$-by-$n$ node similarity matrix; therefore, the major advantage of our proposed method compared to AGCRN is that our method has a smaller memory cost as it does not compute any $n$-by-$n$ matrices.

D$^2$STGNN decouples the signal into two parts with an estimation gate. 
The first part is processed with diffusion convolutional layers and the second part is processed with gated recurrent units (GRU) layers. 
In addition, D$^2$STGNN has a dynamic graph learning model which learns the $n$-by-$n$ dynamic transition matrices that indicate the relationship between different nodes. 
One similarity between D$^2$STGNN and our proposed method is that D$^2$STGNN also has an ensemble-like structure where the decoupled signals are processed with two different models (i.e., the convolutional layers and GRU layers). 
This suggests that an ensemble-like structure could be important for achieving competitive performances on spatial-temporal datasets. 
D$^2$STGNN learns the dependency between different nodes with the diffusion convolutional layers, and our method learns such information with spatial mixers. 
The ability to learn inter-node dependency is crucial for achieving competitive performances. 
The major advantage of our proposed method compared to D$^2$STGNN is that our method has a smaller memory cost as it does not compute any $n$-by-$n$ matrices.

DGCRN is a graph convolutional recurrent network capable of generating dynamic graphs (i.e., $n$-by-$n$ dynamic adjacency matrices to capture inter-node dependency. 
Our method internally learns the inter-node dependency with spatial mixers. 
For both methods, the ability to learn inter-node dependency is crucial for achieving competitive performances. 
The major difference between DGCRN and our method is that DGCRN uses graph convolution and recurrent operations to process the input signal, while our method uses only linear layers to process the signal. 
The major advantage of our proposed method compared to DGCRN is that our method has a smaller memory cost as it does not compute any $n$-by-$n$ matrices.

GWNET consists of several convolutional blocks, each containing two types of layers: 1) gated temporal convolutional layers and 2) diffusion convolutional layers. 
The major difference between the two methods is that GWNET uses convolutional layers to process the signals, whereas our proposed method uses linear layers. 
The first similarity shared by GWNET and our proposed method is that both methods adopt residual layers. 
In other words, the blocks in GWNET could have ensemble-like behavior, and such design may contribute greatly to the success of both methods. 
Another similarity is that D$^2$STGNN learns the dependency between different nodes with the diffusion convolutional layers, and our method learns such information with spatial mixers. 
The ability to learn inter-node dependency is also crucial for achieving competitive performances. 
Because GWNET computes $n$-by-$n$ self-adaptive adjacency matrix in diffusion convolutional layers, the major advantage of our proposed method compared to GWNET is that our method has a smaller memory cost as it does not compute any $n$-by-$n$ matrices.

TSMixer is the method most similar to ours. 
The two differences are the complex linear layers and random projection layers. 
According to the ablation study (see Section~\ref{sec:ablation}), the majority of the performance gain came from the random projection layers.

In conclusion, the capability of learning inter-node dependency is crucial for the success of a method. 
The random projection layer design is the major reason for our model to outperform the baseline method (i.e., TSMixer) that is most similar to ours.

\subsection{Ablation Study}
\label{apdx:ablation}
\begin{figure}[H]
\centerline{
\includegraphics[width=0.99\linewidth]{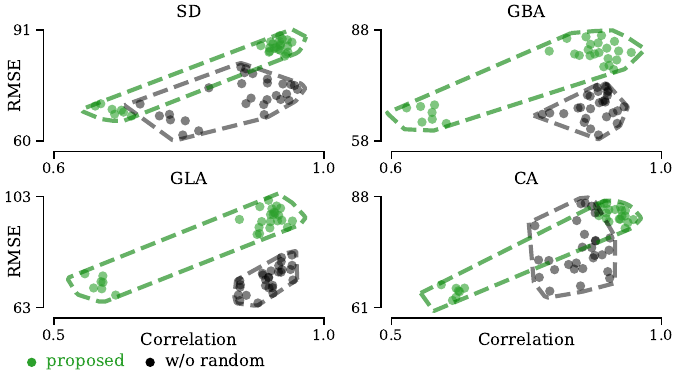}
}
\caption{
The correlation-error diagram illustrates the interplay of diversity and error (RMSE) across individual blocks in the model.
}
\label{fig:corr_error_rmse}
\end{figure}

\begin{figure}[H]
\centerline{
\includegraphics[width=0.99\linewidth]{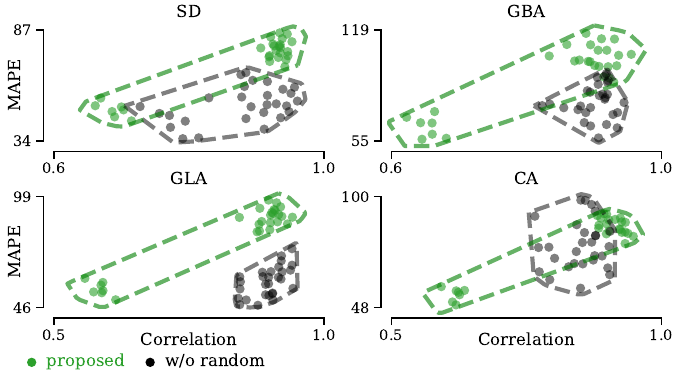}
}
\caption{
The correlation-error diagram illustrates the interplay of diversity and error (MAPE) across individual blocks in the model.
}
\label{fig:corr_error_mape}
\end{figure}

We have produced correlation-error diagrams employing all three measurements (MAE, RMSE, and MAPE). 
In Section~\ref{sec:ablation}, we showcase the correlation-error diagram with MAE in Fig.~\ref{fig:corr_error_mae}. 
This section introduces the remaining two correlation-error diagrams, namely, the correlation-error diagram with RMSE in Fig.~\ref{fig:corr_error_rmse}, and the correlation-error diagram with MAPE in Fig.~\ref{fig:corr_error_mape}. 
The conclusions remain consistent. 
The random projection layer aids the proposed method in achieving greater diversity.

\begin{figure}[htp]
\centerline{
\includegraphics[width=0.99\linewidth]{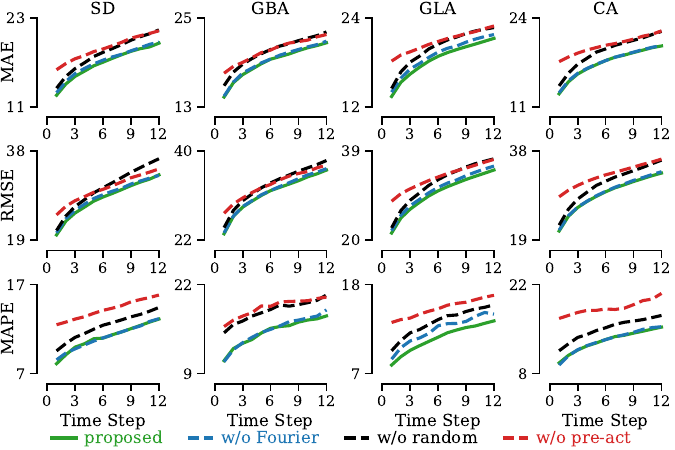}
}
\caption{
The ablation study result measured using per-step MAE, RMSE, and MAPE.
The random projection layer benefits the prediction for later time steps more compared to earlier time steps.
}
\label{fig:ablation_step}
\end{figure}

To understand the effect of different design choices at various time horizons, we visualized the performance of different variants of the proposed model over the prediction time steps in Fig.~\ref{fig:ablation_step}. 
For the pre-activation and frequency domain choices, the positive effect is evenly distributed across different time steps. 
For the random projection layer, the benefits lean more towards longer-term predictions. 
A possible explanation is that the variance for longer-term prediction is typically higher. 
Since the random projection layer helps different mixer blocks focus on different dimensions of the multivariate time series, it aids the model in better capturing the later, harder-to-predict time steps.

\subsection{Runtime and Space Complexity}
The proposed \proposed{}'s efficiency closely aligns with that of the TSMixer, as both employ a similar all-MLP structure. 
With this in mind, we have conducted experiments to compare the runtimes of both the TSMixer and the proposed method. 
Table~\ref{tab:runtime} shows the average runtime for each epoch.

\begin{table}[htp]
\caption{Runtime in seconds.}
\label{tab:runtime}
\begin{center}
\begin{tabular}{lcccc}
\shline
Data                & SD  & GBA & GLA & CA   \\ \hline 
TSMixer             & 196 & 591 & 954 & 2115 \\
RPMixer             & 72  & 179 & 261 & 589  \\
RPMixer w/o Fourier & 46  & 85  & 106 & 231  \\
RPMixer w/o random  & 69  & 168 & 250 & 568  \\ \shline
\end{tabular}
\end{center}
\end{table} 

The TSMixer is slower comparing to \proposed{} due to its incorporation of batch normalization layers. 
When examining the different variants of \proposed{}, we observe that the complex linear layer has a longer runtime compared to the random projection layer.
Regarding the space complexity of the intermediate representation for all methods enumerated in Table~\ref{tab:runtime}, it is $O(nm)$, where $n$ denotes the number of nodes, and $m$ denotes the length of the input time series.

\subsection{Long-Term Time Series Forecasting}
\label{apdx:long_term}
The effectiveness of the proposed method in spatial-temporal forecasting tasks has been demonstrated. 
However, as a general time series forecasting method, we sought to assess its performance against other forecasting methods. 
To this end, we evaluated our method on seven multivariate long-term time series forecasting datasets (i.e., ETTh1, ETTh2, ETTm1, ETTm2, Weather, Electricity, and Traffic), and compared it with two state-of-the-art methods, TSMixer~\cite{chen2023tsmixer} and PatchTST~\cite{nie2022time}, which have shown superior performance over alternatives such as Autoformer~\cite{wu2021autoformer}, Informer~\cite{zhou2021informer}, TFT~\cite{lim2021temporal}, FEDformer~\cite{zhou2022fedformer}, and linear models~\cite{zeng2023transformers}.

The experimental setup followed the guidelines outlined in~\cite{nie2022time,chen2023tsmixer}.
We evaluated the models under long-term time series forecasting settings, with prediction lengths of 96, 192, 336, and 720.
The evaluation metrics employed were mean square error (MSE) and mean absolute error (MAE). 
The dataset was divided into training, validation, and testing subsets, as suggested in~\cite{nie2022time,chen2023tsmixer,zeng2023transformers}.
The input length was set to 512 as per~\cite{chen2023tsmixer}, and other hyper-parameters were determined based on the validation set results.
The settings for the hyper-parameters are provided with the source code, which can be downloaded from~\cite{supplementary}.
We employed the AdamW optimizer~\cite{loshchilov2017decoupled} with the objective of minimizing the mean square error.
The results are presented in Table~\ref{tab:ltsf_benchmark}.
The performance for TSMixer and PatchTST were obtained from their respective papers~\cite{nie2022time,chen2023tsmixer}.

\begin{table*}[tp]
\caption{Performance comparisons. 
We bold the best-performing results.}
\label{tab:ltsf_benchmark}
\begin{center}
\small
\begin{tabular}{llcc|cc|cc|cc}
\shline 
\multirow{2}{*}{Data}        & \multirow{2}{*}{Method} & \multicolumn{2}{c}{Horizon  96} & \multicolumn{2}{c}{Horizon  192} & \multicolumn{2}{c}{Horizon  336} & \multicolumn{2}{c}{Horizon  720} \\ \cline{3-10} 
                             &                         & MSE            & MAE            & MSE             & MAE            & MSE             & MAE            & MSE             & MAE            \\ \hline \hline
\multirow{3}{*}{ETTh1}       & PatchTST                & 0.370          & 0.400          & 0.413           & 0.429          & 0.422           & 0.440          & \textbf{0.447}  & \textbf{0.468} \\
                             & TSMixer                 & \textbf{0.361} & \textbf{0.392} & \textbf{0.404}  & \textbf{0.418} & \textbf{0.420}  & \textbf{0.431} & 0.463           & 0.472          \\
                             & \proposed{}                & 0.444          & 0.444          & 0.488           & 0.475          & 0.521           & 0.498          & 0.625           & 0.574          \\ \hline \hline
\multirow{3}{*}{ETTh2}       & PatchTST                & 0.274          & 0.337          & 0.341           & 0.382          & 0.329           & 0.384          & 0.379           & 0.422          \\
                             & TSMixer                 & 0.274          & 0.341          & 0.339           & 0.385          & 0.361           & 0.406          & 0.445           & 0.470          \\
                             & \proposed{}                & \textbf{0.173} & \textbf{0.284} & \textbf{0.210}  & \textbf{0.317} & \textbf{0.246}  & \textbf{0.346} & \textbf{0.340}  & \textbf{0.410} \\ \hline \hline
\multirow{3}{*}{ETTm1}       & PatchTST                & 0.293          & 0.346          & 0.333           & 0.370          & 0.369           & 0.392          & \textbf{0.416}  & 0.420          \\
                             & TSMixer                 & \textbf{0.285} & \textbf{0.339} & \textbf{0.327}  & \textbf{0.365} & \textbf{0.356}  & \textbf{0.382} & 0.419           & \textbf{0.414} \\
                             & \proposed{}                & 0.358          & 0.385          & 0.397           & 0.406          & 0.439           & 0.433          & 0.503           & 0.475          \\ \hline \hline
\multirow{3}{*}{ETTm2}       & PatchTST                & 0.166          & 0.256          & 0.223           & 0.296          & 0.274           & 0.329          & 0.362           & 0.385          \\
                             & TSMixer                 & 0.163          & 0.252          & 0.216           & 0.290          & 0.268           & 0.324          & 0.420           & 0.422          \\
                             & \proposed{}                & \textbf{0.111} & \textbf{0.224} & \textbf{0.139}  & \textbf{0.252} & \textbf{0.168}  & \textbf{0.276} & \textbf{0.212}  & \textbf{0.311} \\ \hline \hline
\multirow{3}{*}{Weather}     & PatchTST                & 0.149          & \textbf{0.198} & 0.194           & \textbf{0.241} & 0.245           & 0.282          & \textbf{0.314}  & \textbf{0.334} \\
                             & TSMixer                 & \textbf{0.145} & \textbf{0.198} & \textbf{0.191}  & 0.242          & \textbf{0.242}  & \textbf{0.280} & 0.320           & 0.336          \\
                             & \proposed{}                & 0.149          & 0.206          & 0.198           & 0.250          & 0.258           & 0.295          & 0.343           & 0.353          \\ \hline \hline
\multirow{3}{*}{Electricity} & PatchTST                & \textbf{0.129} & \textbf{0.222} & \textbf{0.147}  & \textbf{0.240} & \textbf{0.163}  & \textbf{0.259} & \textbf{0.197}  & \textbf{0.290} \\
                             & TSMixer                 & 0.131          & 0.229          & 0.151           & 0.246          & 0.161           & 0.261          & \textbf{0.197}  & 0.293          \\
                             & \proposed{}                & 0.130          & 0.229          & 0.149           & 0.247          & 0.166           & 0.264          & 0.201           & 0.302          \\ \hline \hline
\multirow{3}{*}{Traffic}     & PatchTST                & \textbf{0.360} & \textbf{0.249} & \textbf{0.379}  & \textbf{0.256} & \textbf{0.392}  & \textbf{0.264} & \textbf{0.432}  & \textbf{0.286} \\
                             & TSMixer                 & 0.376          & 0.264          & 0.397           & 0.277          & 0.413           & 0.290          & 0.444           & 0.306          \\
                             & \proposed{}                & 0.394          & 0.277          & 0.406           & 0.282          & 0.415           & 0.286          & 0.451           & 0.307 \\ \shline    
\end{tabular}
\end{center}
\end{table*}

The results suggest that our proposed method performs comparably to both TSMixer and PatchTST, as confirmed by a $t$-test with $\alpha=0.05$. 
This implies that the proposed method also attains state-of-the-art performance on the long-term time series forecasting tasks. 
Chen et al.~\cite{chen2023tsmixer} noted that the cross-variate information might not be as significant in these seven datasets.  
However, it has been demonstrated in~\cite{liu2023itransformer} that effectively capturing the cross-variate information using an attention mechanism can enhance performance. 
This observation suggests that merging the random projection concept with an attention mechanism could be a promising future direction. 
It should be noted that the dimension count of most long-term time series forecasting datasets is significantly lower than that of large-scale traffic datasets. 
The modest performance of the proposed method for these datasets aligns with the results presented in Table~\ref{tab:largest_benchmark}. 
The dataset with the fewest dimensions exhibits the least improvement compared to the baseline methods.

\subsection{Comparative Analysis of \proposed{} and Alternative Forecasting Methods}
The experimental section primarily compared our proposed \proposed{} method with baseline methods that have reported results on the LargeST dataset~\cite{liu2023largest} available at the time of writing. 
TSMixer~\cite{chen2023tsmixer} is the only deep learning method that deviates from this rule, as it significantly inspired the \proposed{} method. 
This section offers an analysis of alternative large-scale spatial-temporal forecasting methods and their potential integration with \proposed{}.

\subsubsection{Spatial-Temporal Forecasting Model}
The PDFormer~\cite{jiang2023pdformer} model, a spatial-temporal graph neural network, employs a stack of transformer encoders with skip connections to process spatial-temporal data. 
Each encoder comprises three attention modules designed to capture long-range spatial dependencies, short-range spatial dependencies, and temporal information within a node. 
The random projection concept from our paper could be integrated into PDFormer by adding a fourth module, the spatial mixer, to the transformer encoder to diversify the intermediate representation output. 
MegaCRN~\cite{jiang2023spatio}, based on a graph convolutional recurrent unit (GCRU) encoder-decoder architecture, uses a meta-graph learner to generate a meta-graph for the recurrent unit. 
As MegaCRN is GCRU-based, it would be challenging to adopt a random projection layer to enhance its performance. 
Spatial and temporal identities are learnable embedding features~\cite{shao2022spatial} that could be combined with different model architectures. 
Thus, it would be interesting to explore how they could further enhance \proposed{}'s performance in large-scale spatial-temporal forecasting.

\subsubsection{General Forecasting Model}
Two significant components proposed in~\cite{li2023revisiting}, reversible normalization (RevIN) and channel independence (CI), aim to address linear layers' limitations. 
RevIN assists linear models in trend prediction, and CI aids linear models in handling multivariate time series with varying periods across different dimensions (or channels). 
RevIN could be incorporated into \proposed{} to boost its trend prediction capability, while CI may not be compatible with \proposed{}, which assumes dimensional dependency. 
The choice between CI and \proposed{} should be based on the dataset's characteristics. 
Factorized multilayer perceptrons (MLPs), proposed in~\cite{li2023mts} and evaluated on common multivariate time series forecasting benchmark datasets, model the dimension and temporal interaction of multivariate time series. 
It would be interesting to evaluate \proposed{}'s performance after substituting the temporal mixer with a factorized MLP on large-scale spatial-temporal datasets. 

The frequency-domain MLP~\cite{yi2024frequency} is a model that processes input multivariate time series with MLP in the frequency domain, using a frequency channel learner and a frequency temporal learner, and does not incorporate mixer blocks. 
The frequency temporal learner could be used as the alternative design for the temporal mixer used in \proposed{}. 
While the FFT design is not our primary contribution, it offers an interesting perspective to assess if replacing FFT with decomposition~\cite{zhou2022fedformer} or downsampling~\cite{zhang2022less} could enhance the handling of periodic data. 
In future work, we could weave these methods into \proposed{} in numerous ways, either incorporating them into every temporal mixer or into the network's input or output.

The N-BEATS~\cite{oreshkin2019n} model is a univariate time series forecasting model composed of a stack of residual building blocks, each a MLP with a forecast output and a backcast output. 
As a univariate time series model, its design principles could be used to refine the temporal model design in \proposed{}.

\subsection{Additional Benchmark Result}
In this section, we present additional experimental results comparing the proposed \proposed{} with another MLP model, Spatial and Temporal IDentity information (STID)~\cite{shao2022spatial}, on LargeST datasets~\cite{liu2023largest}. 
We predominantly adopted the hyper-parameter settings from~\cite{shao2022spatial, shao2022spatial2}. 
The experimental results, averaged over the full 12 time horizon, are displayed in Table~\ref{tab:stid_benchmark}.

\begin{table}[htp]
\caption{Comparisons between \proposed{} and STID.}
\label{tab:stid_benchmark}
\begin{center}
\begin{tabular}{llcccc}
\shline
Data                 & Method  & Param & MAE   & RMSE  & MAPE  \\ \hline \hline
\multirow{2}{*}{SD}  & \proposed{} & 1.5M  & 16.90 & 27.97 & 11.07 \\
                     & STID    & 127K  & 29.05 & 47.73 & 19.28 \\ \hline \hline
\multirow{2}{*}{GBA} & \proposed{} & 2.3M  & 19.06 & 31.54 & 15.09 \\
                     & STID    & 180K  & 31.20 & 48.72 & 26.11 \\ \hline \hline
\multirow{2}{*}{GLA} & \proposed{} & 3.2M  & 18.46 & 30.13 & 11.34 \\
                     & STID    & 227K  & 31.86 & 50.51 & 20.20 \\ \hline \hline
\multirow{2}{*}{CA}  & \proposed{} & 7.8M  & 17.50 & 28.90 & 13.03 \\
                     & STID    & 380K  & 29.18 & 46.78 & 21.40 \\ \shline
\end{tabular}
\end{center}
\end{table} 

Overall, \proposed{} surpasses STID in all error measurements. 
In terms of runtime, STID's duration is roughly half that of \proposed{}, due to STID's smaller model size. 
Nevertheless, STID's key innovation, i.e., learnable spatial and temporal embeddings~\cite{shao2022spatial}, is versatile and can be integrated with \proposed{}. 
We performed an ablation study to assess the impact of the STID components. 
The results are available in Table~\ref{tab:stid_ablation}.

\begin{table}[htp]
\caption{Ablation study of \proposed{} with STID component.}
\label{tab:stid_ablation}
\begin{center}
\resizebox{\linewidth}{!}{%
\begin{tabular}{llcccc}
\shline
Data                 & Method  & Param & MAE   & RMSE  & MAPE  \\ \hline \hline
\multirow{3}{*}{SD}  & \proposed{}                           & 1.5M  & \textbf{16.90}  & 27.97 & 11.07 \\
                     & ~+ STID               & 2.6M  & 16.91 & \textbf{27.72} & \textbf{11.04} \\
                     & ~+ STID - random - Fourier & 1.6M  & 17.71 & 28.64 & 11.98 \\ \hline \hline
\multirow{3}{*}{GBA} & \proposed{}                            & 2.3M  & \textbf{19.06} & 31.54 & \textbf{15.09} \\
                     & ~+ STID                    & 3.6M  & 19.07 & \textbf{31.02} & 15.74 \\
                     & ~+ STID - random - Fourier & 3.3M  & 20.58 & 32.45 & 18.24 \\ \hline \hline
\multirow{3}{*}{GLA} & \proposed{}                            & 3.2M  & \textbf{18.46} & 30.13 & \textbf{11.34} \\
                     & ~+ STID                    & 4.7M  & 18.73 & \textbf{30.06} & 11.75 \\
                     & ~+ STID - random - Fourier & 5.4M  & 20.04 & 31.62 & 13.44 \\ \hline \hline
\multirow{3}{*}{CA}  & \proposed{}                            & 7.8M  & \textbf{17.50}  & 28.90  & \textbf{13.03} \\
                     & ~+ STID                    & 9.7M  & 17.55 & \textbf{28.74} & 13.22 \\
                     & ~+ STID - random - Fourier & 14.9M & 18.91 & 30.05 & 16.09 \\ \shline
\end{tabular}
}
\end{center}
\end{table} 

In this ablation study, we considered three settings: 1) \proposed{}, the method introduced in this paper, 2) \proposed{} + STID, which is \proposed{} enhanced with the learnable spatial and temporal embeddings from STID, and 3) \proposed{} + STID - Random - Fourier, the previous setting with the main components (i.e., random projection layers and complex linear layers) of \proposed{} turned off. 
The first setting serves as the baseline for this study, the second setting demonstrates the potential benefit of the STID components on the proposed method, and the third setting illustrates the relative importance of the STID components versus the \proposed{} components. 
By integrating STID into \proposed{} (i.e., \proposed{} + STID), the RMSE on all datasets improves. 
However, when the \proposed{} components are turned off (i.e., \proposed{} + STID - Random - Fourier), the performance on all three measures deteriorates. 
The \proposed{} components have a greater impact on performance compared to the STID components.

The number of parameters for \proposed{} + STID is higher compared to \proposed{}, due to the inclusion of additional embedding layers. 
In terms of runtime, incorporating STID increases the runtime by an average of 22\% because of these extra layers. 
Overall, integrating STID components into \proposed{} could be a promising strategy to further enhance the performance of \proposed{}.

\subsection{Interpretability and Ease of Use}
Regarding interpretability, \proposed{} does not compute adjacency matrices like many baseline methods, thus it does not provide intrinsic interpretability. 
However, post-hoc interpretability can be achieved through methods like sensitivity analysis~\cite{wang2019deepvid,ismail2019deep,wang2021visual,yeh2023multitask}, which can highlight the more important parts of the input. 
In terms of ease of use, \proposed{} is composed of four common layers: linear, \textsc{ReLU}, FFT, and iFFT, which can be found in almost all deep learning libraries. 
Therefore, we believe \proposed{} is relatively easy to implement.

\end{document}